\documentclass[conference]{IEEEtran}
\IEEEoverridecommandlockouts
\usepackage{cite}
\usepackage{amsmath,amssymb,amsfonts}
\usepackage{algorithmic}
\usepackage{graphicx}
\usepackage{textcomp}
\usepackage{xcolor}
\usepackage{caption}
\usepackage{url}
\usepackage{subcaption}
\usepackage{graphicx}
\usepackage{wrapfig}
\def\BibTeX{{\rm B\kern-.05em{\sc i\kern-.025em b}\kern-.08emT\kern-.1667em\lower.7ex\hbox{E}\kern-.125emX}}
\begin{document}
\title{Language-Agnostic Analysis of Speech Depression Detection\\
}
\author{
\IEEEauthorblockN{1\textsuperscript{st} Sona Binu}
\IEEEauthorblockA{\textit{CSE, Saintgits College of Engineering }\\
\textit{Kottayam, India}}
\and
\IEEEauthorblockN{2\textsuperscript{nd} Jismi Jose}
\IEEEauthorblockA{\textit{CSE, Saintgits College of Engineering }\\
\textit{Kottayam, India}}
\and
\IEEEauthorblockN{3\textsuperscript{rd} Fathima Shimna K V}
\IEEEauthorblockA{\textit{CSE, Saintgits College of Engineering }\\
\textit{Kottayam, India}}
\and
\IEEEauthorblockN{4\textsuperscript{th} Alino Luke Hans}
\IEEEauthorblockA{\textit{CSE, Saintgits College of Engineering }\\
\textit{Kottayam, India}}
\and
\IEEEauthorblockN{5\textsuperscript{th} Reni K. Cherian}
\IEEEauthorblockA{\textit{CSE, Saintgits College of Engineering}\\
reni.cherian@saintgits.org}
\and
\IEEEauthorblockN{6\textsuperscript{th} Starlet Ben Alex}
\IEEEauthorblockA{\textit{ECE, Saintgits College of Engineering}\\
starlet.ben@saintgits.org}
\and
\IEEEauthorblockN{7\textsuperscript{th} Priyanka Srivastava}
\IEEEauthorblockA{\textit{Cognitive Science Lab, IIIT Hyderabad, India} \\
priyanka.srivastava@iiit.ac.in}
\and
\IEEEauthorblockN{8\textsuperscript{th} Chiranjeevi Yarra}
\IEEEauthorblockA{\textit{Language Technologies Research Center, IIIT Hyderabad, India}\\
chiranjeevi.yarra@iiit.ac.in}         
}

\maketitle
\begin{abstract}
The people with Major Depressive Disorder (MDD) exhibit the symptoms of tonal variations in their speech compared to the healthy counterparts. However, these tonal variations not only confine to the state of MDD but also on the language, which has unique tonal patterns. This work analyzes automatic speech-based depression detection across two languages, English and Malayalam, which exhibits distinctive prosodic and phonemic characteristics. We propose an approach that utilizes speech data collected along with self-reported labels from participants reading sentences from IViE corpus, in both English and Malayalam. The IViE corpus consists of five sets of sentences: simple sentences, WH-questions, questions without morphosyntactic markers, inversion questions and coordinations, that can naturally prompt speakers to speak in different tonal patterns. Convolutional Neural Networks (CNNs) are employed for detecting depression from speech. The CNN model is trained to identify acoustic features associated with depression in speech, focusing on both languages. The model's performance is evaluated on the collected dataset containing recordings from both depressed and non-depressed speakers, analyzing its effectiveness in detecting depression across the two languages. Our findings and collected data could contribute to the development of language-agnostic speech-based depression detection systems, thereby enhancing accessibility for diverse populations.
\end{abstract}


\section{Introduction}
Depression is a common mental health disorder characterized by continuous feelings of despair, sadness and a lack of interest or enjoyment in activities for a long period of time. It affects people of all ages, races and socioeconomic backgrounds. Depression is different from normal mood changes and everyday feelings. It can impact all areas of life, including relationships with family, peers and the community. Depression can cause problems at school and work or it can result from those. About 3.8\% of the world’s population faces depression. This includes 5\% of adults, with 4\% of men and 6\% of women affected. Around 5.7\% of adults over 60 also suffer from depression. In total, about 280 million people worldwide suffer from depression. Women are about 50\% more likely to be depressed than men. Over 10\% of pregnant women and new mothers worldwide experience depression. Each year, over 700,000 people die by suicide. It is the fourth most common cause of death for people aged 15 to 29.

At its core, depression changes not only one's emotional topography but also vary the rhythms of communication. Prosody, which includes the tone, rhythm, and flow of language, is a key part of all natural languages. It can provide structure and sometimes change the meaning of a sentence. Although prosodic features of depression symptoms can be observed not only in spoken languages but also in the hand gestures, facial expressions, head movements and body language of sign communication, we consider prosodic features in spoken language to be crucial for this study. In spoken language, prosody shows up as pitch, speed, volume and pauses. Prosody helps to incorporate verbal expression into one's speech, for example, changing the prosody, especially the tone of a sentence, can change the whole meaning of what is said. In English, if a sentence ends with a high pitch, it can sound like a question, even if it's not actually a question (like saying 'you're coming'). On the otherhand, prosody is also used to express deeper emotions in speech production and to perceive the emotions behind one's speech.

Communication is not just about words; how things are said is also important. This is where prosody comes into play. Emotional prosody is recognized as a sophisticated aspect of communication that complements the words used to express deeper emotions. It involves both speech patterns and the ability to discern the emotions behind someone’s speech. By exploring emotional prosody, insights into how language and emotions interact during communication can be gained. The mind and body are so interconnected that voice problems are closely linked to stress and depression, either as primary or secondary factors. Also, a significant correlation between prosody and factors such as turn-taking, reciprocity, interpersonal outcomes and the assessment of depression has been indicated by research. Over the past 20 years, studies have focused on the prosody and acoustic characteristics of individuals with depressive disorders (PWDD). It has been demonstrated by Yang et al. that prosody analysis is a valuable approach for assessing and illustrating the severity of depression. Prosody includes components such as intensity, fundamental frequency (F0), time (interpreted as speech rate) and other related characteristics like formants, jitter, shimmer, and cepstrum \cite{b4}. Two prosodic elements derived from responses to the first three questions on the Hamilton Rating Scale for Depression (HRS-D) that assess primary depression symptoms such as suicidal thoughts, feelings of guilt and low mood, were analyzed \cite{b5}.

Understanding how emotions and verbal expression are influenced by prosody is considered crucial for sorting out the complexities of human communication. By exploring both aspects, valuable insights can be gained into how cues related to depression are conveyed through verbal communication in different languages. To address this, a dataset was created by selecting 22 English sentences from the IViE corpus \cite{b14}, translating them into Malayalam, and producing a total of 44 sentences across five categories. Audio samples were collected from 132 participants, each of whom provided 44 recordings. Before the recordings, surveys were filled out by participants to assess depression, anxiety, and mood. The collected audio was then preprocessed using techniques like noise addition and pitch shifting, and features such as Zero Crossing Rate and MFCC were extracted. Following this, a 1D CNN model with four convolutional layers was developed, utilizing ReLU activation, max-pooling and a fully connected layer. The model was trained in batches of 64 samples over 50 epochs, with a reduced learning rate used to enhance performance. After training, an impressive accuracy of 76\% was achieved on the test dataset, demonstrating the model’s ability to accurately classify emotional states from audio.

\section{Motivation}
Depression is recognized as a prevalent mental disorder associated with significant impairment in mental health and is commonly comorbid with other neuropsychiatric disorders \cite{b1}. The significance of using real clinical data for depression detection research is emphasized by the Chinese clinical depression corpus collected from HAMD interviews at Beijing Huilongguan Hospital \cite{b6}. Various algorithms have been considered for classification, including Convolutional Neural Networks (CNN), Long Short-Term Memory (LSTM) and Time-Delay Neural Network (TDNN) \cite{b7}. The Multimodal Speech Chain Depression Recognition (MSCDR) model, which extracts speech sounds to automatically detect depression based on how people speak, is discussed in \cite{b12}. The importance of features such as pitch (fundamental frequency, Fo), energy and speaking rate in capturing emotional variations in speech is highlighted by Moore et al. \cite{b3}. By examining these prosodic features at the sentence level and grouping them into observations, the study aims to quantify emotional expressions and variations in speech patterns among individuals with depressive disorders \cite{b2}.

Most existing works have either involved in the development of monolingual speech corpora for depression detection or the construction of models for detecting depression using monolingual speech. But no analysis has been conducted on the impact of prosody aspects related to verbal communication on depression detection. Also, no publicly available dataset exists that is collected from a depressed population speaking prompts that implicitly trigger verbal expression related prosody. Besides, no corpus facilitates the study of variations in depression cues across multiple languages spoken by the same speakers. Symptoms of depression may vary across cultures and symptoms observed in an English speaker may not directly translate to a non-English speaker. Thus, this could be useful for detecting depression in a bilingual or multilingual society like India. Although various datasets containing speech data for depression research exist, there is a lack of Indian languages, such as Malayalam.

\section{Dataset creation}
A speech corpus was recorded from bilingual Malayalam speakers using stimuli chosen from the IViE corpus, which includes a variety of English dialects and sentence types. This approach enabled prosodic variations to be captured in both English and Malayalam, ensuring a comprehensive dataset of 44 sentences per participant in each language.
\paragraph{Stimuli selection} The stimuli for English recordings were chosen from the existing IViE corpus \cite{b14}, which contains recordings of speakers from various regions in the British Isles and Ireland, offering a range of English dialects. A total of 22 sentences are included in the corpus, comprising eight simple sentences, three questions without morphosyntactic markers, three inversion questions, three WH-questions and five coordinations. The IViE corpus was developed to analyze prosodic variations in verbal communication across different dialects. The stimuli were selected carefully to naturally embed prosodic variations in the verbal expressions of native English speakers. Sentences from the IViE corpus were utilized to create a dataset for the proposed analysis. For the recording of English speech corpora from bilingual Malayalam speakers, the English prompts were used directly. For the recording of Malayalam speech, the English prompts were translated into Malayalam. These sentences were read aloud by participants in both English and Malayalam, capturing speech samples across distinct languages. This process ensured that our dataset incorporated the prosodic variations inherent to each language.

\paragraph{Recording} The speech corpus consisted of data from a total of 132 participants (male and female) who were proficient in both English and Malayalam. Participants were aged between 18 and 35. Participation was entirely voluntary, and no compensation was provided to the participants. All recordings were collected in a room with sufficient acoustics to ensure relatively clean audio samples. To safeguard the privacy of the participants, sensitive personal information was removed before further analysis. A total of 44 sentences (22 in English and 22 in Malayalam) were recorded from each participant.

\subsection{Depression state annotations}
We obtain the participants' self-reported depression state from all 132 participants using the following measures: 1. Patient Health Questionnaire-9 (PHQ-9), 2. Positive and Negative Affect Schedule (PANAS), 3. General Anxiety Disorder (GAD) and State-Trait Inventory (STAI-T). Details of each process are provided in the following subsections.

\subsubsection{PHQ-9}\label{AA}
The PHQ-9 is a widely used screening tool for assessing the severity of depressive symptoms in individuals, according to the guidelines for major depressive disorder in the Diagnostic and Statistical Manual of Mental Disorders (DSM-5) \cite{b8}. It consists of nine questions covering various symptoms of depression, including depressed mood, loss of interest or enjoyment, changes in eating or weight, sleep problems, tiredness, feelings of worthlessness or guilt and trouble concentrating, psychomotor agitation or retardation and thoughts of death or suicide.

Each question on the PHQ-9 is scored from 0 to 3, indicating "not at all," "several days," "more than half the days," and "nearly every day" experienced over the last two weeks. The score ranges from 0 to 27, with higher scores showing greater severity of depressive symptoms. Participants are classified into five categories based on their total scores: 1. None (0-4), 2. Mild depression (5-9), 3. Moderate depression (10-14), 4. Moderately severe depression (15-19), and 5. Severe depression (20-27). Based on the collected scores, the distribution of participants across these categories is as follows: 46 (34.85\%), 52 (39.39\%), 23 (17.42\%), 11 (8.33\%) and 0 (0\%), respectively.

\subsubsection{PANAS}
The PANAS is a widely used self-report questionnaire designed to measure two broad mood dimensions: Positive Affect (PA) and Negative Affect (NA). PA refers to the extent to which an individual experiences positive emotions such as joy, excitement and enthusiasm, while NA refers to the extent of experiencing negative emotions such as sadness, anxiety and anger. Each participant was asked to take the PANAS test prior to their recording to assess their current mood. It is observed that a total of 104 (78.78\%), 17 (12.87\%) and 11 (8.33\%) found to be positive, negative and neutral respectively. 

\subsubsection{GAD}\label{SCM}
The GAD-7 is a self-report questionnaire commonly used to screen people for Generalized Anxiety Disorder (GAD) and to measure the severity of symptoms. \cite{b13} It consists of a total of seven questions that inquire about the frequency and impact of various anxiety-related experiences over the past two weeks. Each question is rated from 0 to 3. The total GAD score ranges from 0 to 21, with scores classified as follows: 0-4 (minimal anxiety), 5-9 (mild anxiety), 10-14 (moderate anxiety) and 15-21 (severe anxiety). The distribution of participants across these categories is as follows: 50 participants (37.8\%), 51 participants (38.6\%), 20 participants (15.15\%) and 11 participants (8.3\%) in the minimal, mild, moderate and severe anxiety categories, respectively.

\subsubsection{STAI-T:}
An individual’s trait anxiety or enduring anxiety is measured using the State-Trait Anxiety Inventory (STAI-T), which indicates their propensity to experience anxiety over time. The STAI-T is a commonly used measure of baseline anxiety tendencies that is beneficial in research and clinical contexts. It is observed that 13 participants (9.84\%) scored below 38, 32 participants (24.24\%) scored between 38 and 44, and 87 participants (65.90\%) scored above 44 on the STAI-T scale.

\section{Experiments}
The experiments are conducted to detect depression considering both English and Malayalam speech corpora using CNNs and features based on depression cues. In order to obtain large dataset for better modelling on CNN, we augmented the data prior to the experiments.  The following sub-sections describe the data augmentation, feature extraction, and architectural details.

\subsection{Data augmentation }
The augmentation process mimics real-world scenarios and introduces variability in speech signals, including background noise, temporal distortions and pitch variations. By exposing the model to a wider range of acoustic conditions, data augmentation mitigates the risk of overfitting and enables the model to learn more robust and discriminative features, thus improving its performance. We have used the following data augmentation strategies.
    \paragraph{Noise Addition} This technique introduces controlled amounts of random noise to the audio data. We simulate real-world environmental conditions and enhance the robustness of the model against noise interference commonly encountered in practical scenarios.
    \paragraph{Time scale modification} This technique alters the duration of the audio signal while preserving its pitch. By applying a stretching factor to the temporal domain, the audio can be simulated to mimic the real-world utterance length variations for the same text prompt.
    \paragraph{Time Shifting} Time shifting involves shifting the entire audio signal along the time axis, introducing temporal displacements without altering the signal's content. By randomly shifting the signal within a specified range, we simulate temporal variations that occur naturally in audio recordings.
    \paragraph{Pitch Shifting} Pitch shifting changes the pitch of the audio signal while preserving its temporal characteristics. By applying pitch shifts within a controlled range, we augment the dataset to include speaker variations with pitch changes, enabling the model to learn robust representations across different speakers.

\subsection{Feature Extraction}
We compute the following features using librosa tool kit \cite{b9}: 1. Zero crossing rate (1 dimension sequence), 2.Chroma (12 dimension feature sequence) \cite{b10}, 3. Mel-frequency cepstral coefficients (MFCC) (36 dimension feature sequence), Short-time energy (1 dimension feature sequence), Mel spectrogram (128 dimension feature sequence). We extract the features for recorded and augmented speech data. We concatenate all the features and obtain 178 dimension feature sequence and given to the model architecture.

\subsection{Architecture}
\begin{figure}[!htbp]
  \centering
  \includegraphics[width=0.5\textwidth, height=0.4\textheight, keepaspectratio]{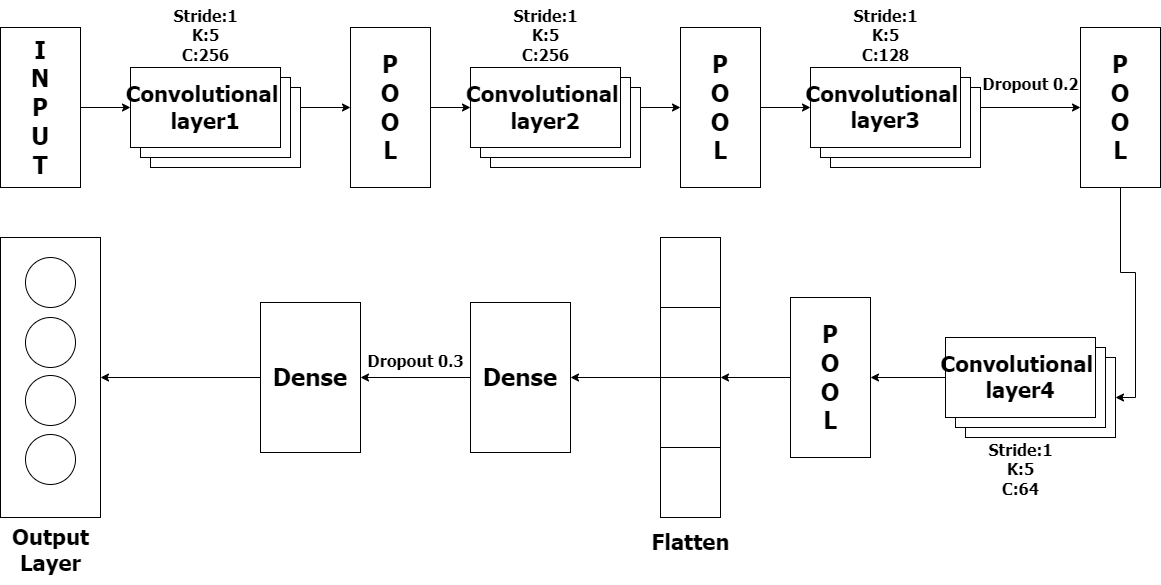}
  \caption{CNN Model}
  \label{fig1}
\end{figure}

The architecture considered for the experimentation are shown in Fig. \ref{fig1}. We considered four convolutional layers. The first layer consists of 256 filters, each with a kernel size of 5. The ReLU (Rectified Linear Unit) activation function is applied to introduce non-linearity. This is followed by a max-pooling layer which reduces the dimensionality of the feature maps by taking the maximum value over a pool size of 5 with a stride of 2. The second layer also has 256 filters and uses the ReLU activation function, followed by max-pooling with a pool size of 5 with a stride of 2. The third layer has 128 filters with a kernel size of 5 and ReLU activation followed by max-pooling with a pool size of 5 with a stride of 2. This layer includes an additional dropout layer with a  20\% dropout rate to mitigate overfitting by randomly dropping neurons during training. The fourth layer has 64 filters with a kernel size of 5, ReLU activation and max-pooling with a pool size of 5 and a stride of 2.

We utilize max-pooling layer after each convolutional layer for utilizing to downsample the feature maps generated by the convolutional layers, retaining the most salient features while reducing spatial dimensions. With a pool size of 5 and strides of 2, each max-pooling operation selects the maximum value within a fixed-size window, effectively reducing the size of the feature maps. By facilitating spatial subsampling and feature extraction, max-pooling layers contribute to the network's ability to learn hierarchical representations of audio data and improve classification performance. In this context, the use of max-pooling aids in reducing computational complexity, promoting translation invariance, and enhancing the network's capacity to detect relevant patterns in the input audio features.\cite{b11}

At the end of fourth convolutaional layer, the output is flattened into a single vector of dimension 704 and fed in to a fully connected layer with 32 hidden units and ReLU activation. This is followed by a dropout layer with a 30\% dropout rate and a softmax activation layer. The architecture is trained with a batch size of 64 over 50 epochs, with a learning rate of 0.4. The data is split as follows: 64\% for training, 16\% for validation and to tune the model's parameters, and 20\% for testing after training is complete. We evaluate the performance of the depression detection based on accuracy and confusion among the classes.

\section{Preliminary Results and analysis}
The Convolutional Neural Network (CNN) model proposed for depression classification based on audio features yielded compelling results in this study. Depression labels were determined based on PHQ-9 ratings, where participants with 'No' or 'Mild' depression were categorized as non-depressed, and those with other ratings were categorized as depressed. Through rigorous training and evaluation, the model achieved an impressive accuracy of 76\% on the test dataset containing both Malayalam and English speech samples. This high level of accuracy indicates the model's proficiency in accurately discerning depression states from audio recordings, a crucial capability for effective depression assessment. Also, an in-depth analysis of the confusion matrix revealed minimal misclassifications across different depression classes, underscoring the model's robustness and its ability to effectively distinguish between various depression states, including mild, moderate, moderately severe and none.

\begin{figure}[!htbp]
  \centering
  \includegraphics[width=0.5\textwidth]{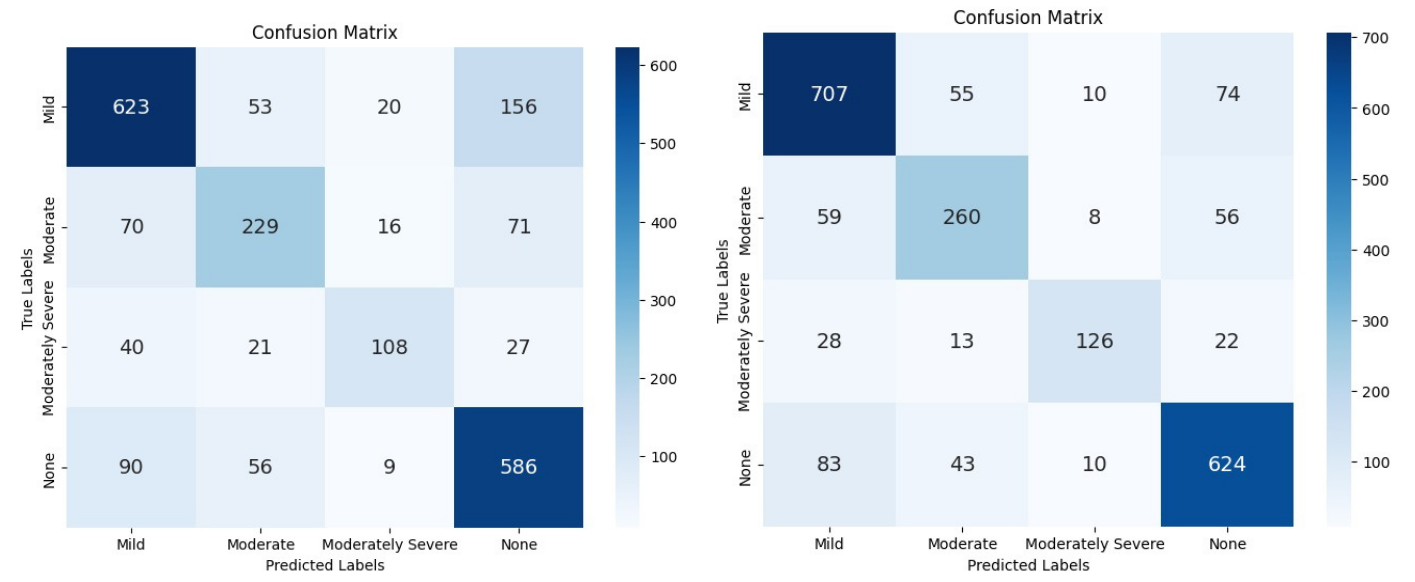}
  \caption{Confusion matrix for English and Malayalam dataset}
  \label{figCME}
\end{figure}

Fig. \ref{figCME} shows the confusion matrix for the four depression states, separately for English and Malayalam speech data. Unlike the overall accuracy presented earlier, this analysis provides a language-agnostic view by examining English and Malayalam separately through confusion matrices. The confusion matrices offer more detailed insights into each depression state compared to a single accuracy measure across all states or classes. From the figures, it is observed that the diagonal entries have higher counts compared to the off-diagonal entries for both Malayalam and English. This indicates that the chosen depression detection model performs consistently across different languages.

\begin{figure}[!htbp]
  \centering
  \includegraphics[width=0.5\textwidth]{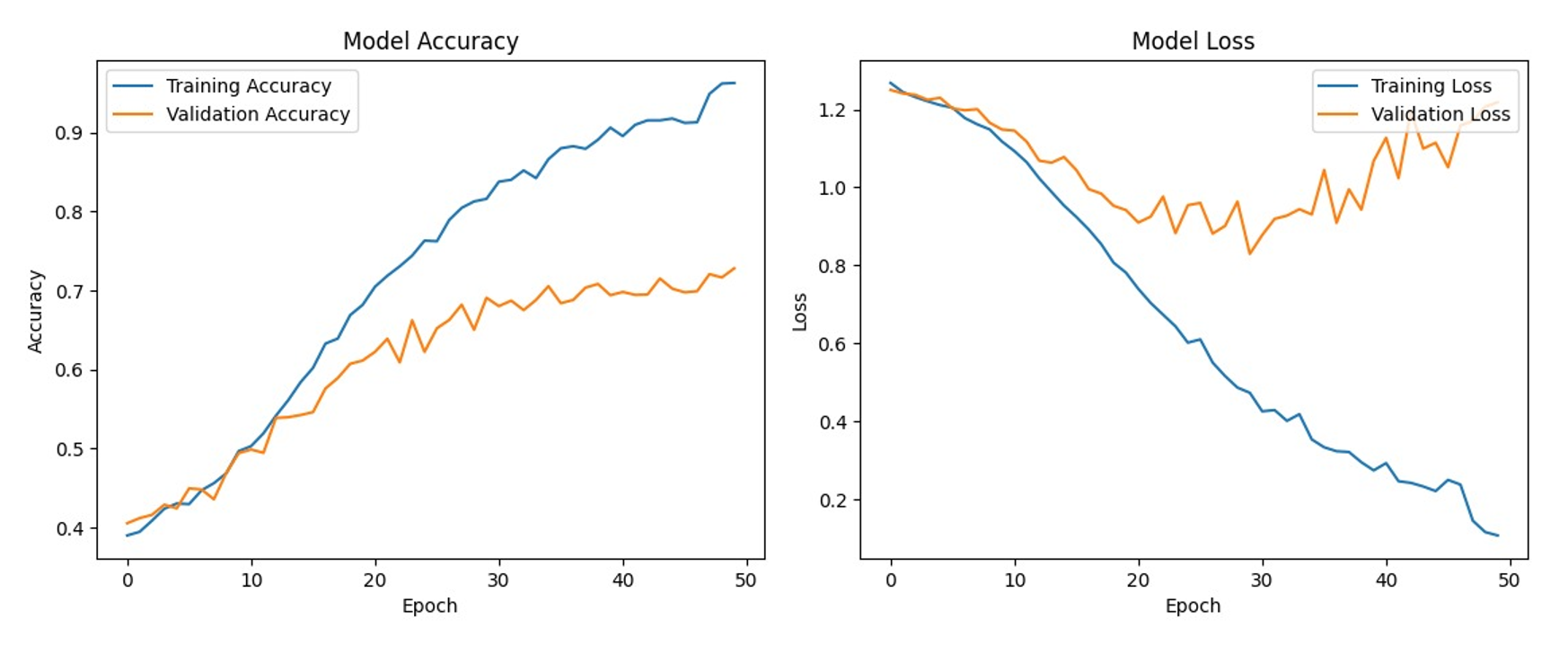}
  \caption{ CNN Model accuracy and loss for English Dataset }
  \label{figCMALED}
\end{figure}

\begin{figure}[!htbp]
  \centering
  \includegraphics[width=0.5\textwidth]{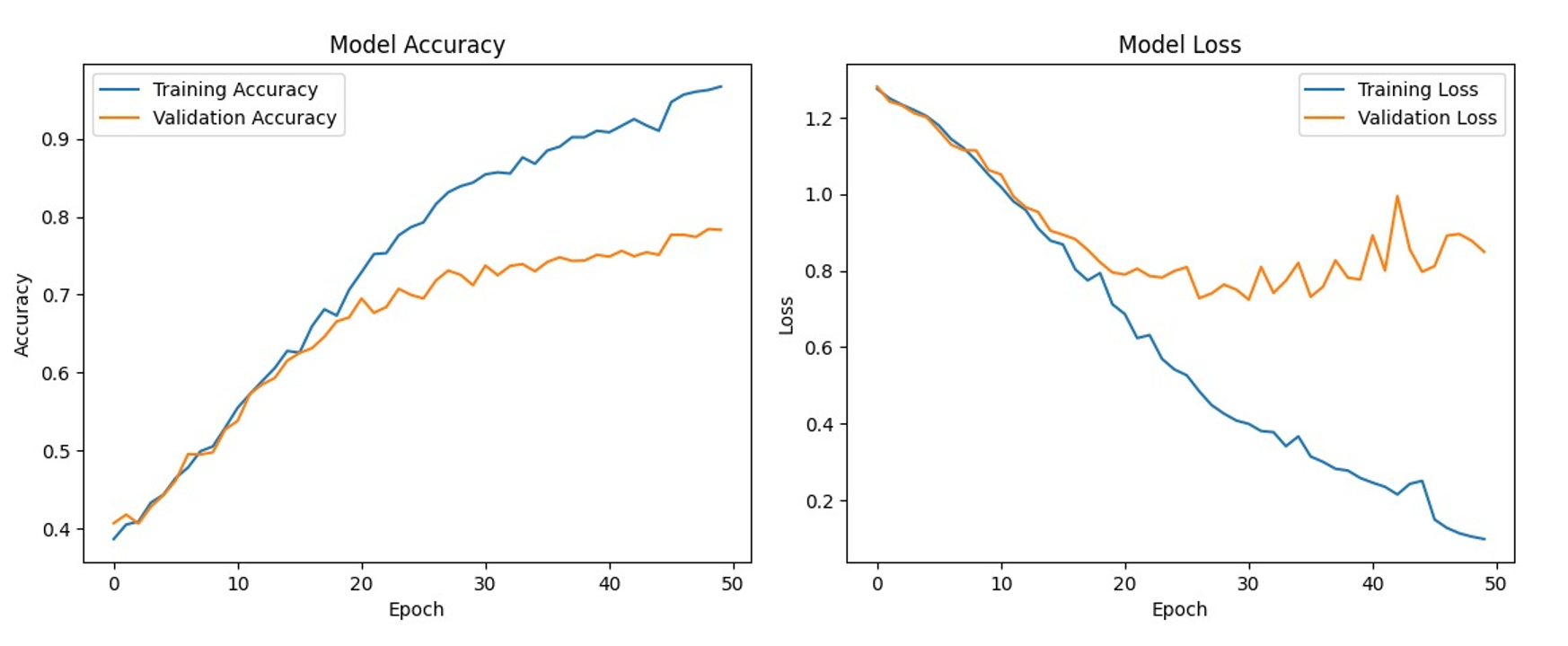}
  \caption{ CNN Model accuracy and loss for Malayalam Dataset }
  \label{figCMALMD}
\end{figure}

\begin{figure}[!htbp]
  \centering
  \includegraphics[width=0.5\textwidth]{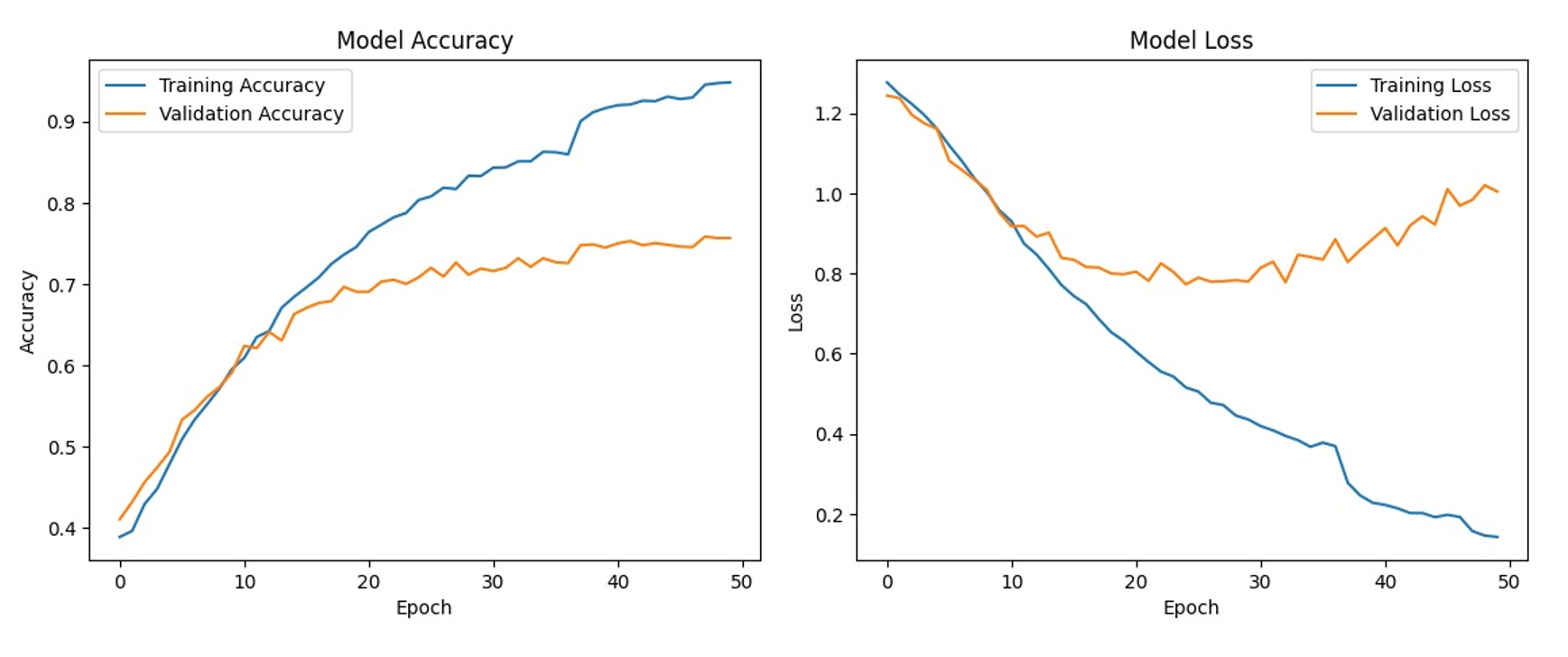}
  \caption{ CNN Model accuracy and loss for Entire Dataset }
  \label{figCMALEMD}
\end{figure}

We further analyzed these results by examining the accuracy and loss curves for CNN model on both English (Fig. \ref{figCMALED}) and Malayalam (Fig. \ref{figCMALMD}) datasets separately as well as on the entire dataset (Fig. \ref{figCMALEMD}). The x-axis represents the training epochs (number of cycles the model trains on the entire dataset), while the y-axis represents the corresponding loss or accuracy values.

The success of the proposed CNN model can be attributed to several factors. Firstly, the utilization of data augmentation techniques, such as noise addition, stretching, and pitching, enriches the training dataset by exposing the model to a diverse range of audio samples. This augmentation enhances the model's ability to work well with new data and improves its robustness against noise and variability inherent in real-world audio recordings. Also, the hierarchical architecture of the CNN allows for automatic feature learning, enabling the model to extract relevant patterns and relationships from the input audio features.

The balanced performance of the model across different depressed classes indicates its capability to handle a variety of depression related expressions effectively. This versatility is particularly advantageous in real-world scenarios where depressed states may vary widely among individuals. Moreover, the model's high accuracy and robustness hold significant implications for clinical practice and mental health research. By automating the process of depression assessment based on audio features, the proposed model offers a non-invasive and efficient means of evaluating mental well-being.

\section{Discussion}
Moving forward, future research endeavors could explore further refinements to the model architecture and the incorporation of additional features to enhance its performance and generalizability. Also, validation studies involving larger and more diverse datasets would give useful information about the model's effectiveness across different demographic groups and cultural contexts. Overall, the results of this study underscore the potential of the proposed CNN model as a valuable tool for automated depression assessment, with promising applications in clinical settings and mental health interventions.

\paragraph{Across four test on annotations} In this study, we conducted a thorough assessment of participants' mental health using standardized tests such as the PHQ-9, STAI-T, PANAS GEN, and GAD scale. Surprisingly, while the PHQ-9 scores exhibited significant correlations with depression levels, the results of the other administered tests failed to demonstrate a statistically significant impact on participants' PHQ-9 scores or depression analysis outcomes. Despite a thorough analysis, the data did not reveal a strong link between the participants' levels of anxiety, as measured by the STAI-T, and their depression severity, as assessed by the PHQ-9. These findings suggest that within our dataset, anxiety levels may not significantly influence depression detection, highlighting the unique sensitivity of the PHQ-9 in capturing depressive symptoms. But it is important to consider that anxiety levels might significantly affect depression in other datasets that are more focused on mental health and designed with relevant sentences or conversations. More research with a larger and more diverse sample may provide additional insights into the subtle relationships between anxiety and depression. However, our dataset, which contains sentences from the IViE Corpus, did not reveal these insights.

\paragraph{Prosody aspects of verbal expression} In this study, we also aimed to analyze the similarities and differences in the perception of language variability in their verbal expressions across English and Malayalam speech. In our study, participants were presented with sentences across five categories: simple sentences, WH-questions, questions without morphosyntactic errors, inversion questions and coordinations. Each participant read the sentences in both English and their translated Malayalam versions. 

\begin{figure}[!htbp]
  \centering
  \includegraphics[width=0.5\textwidth]{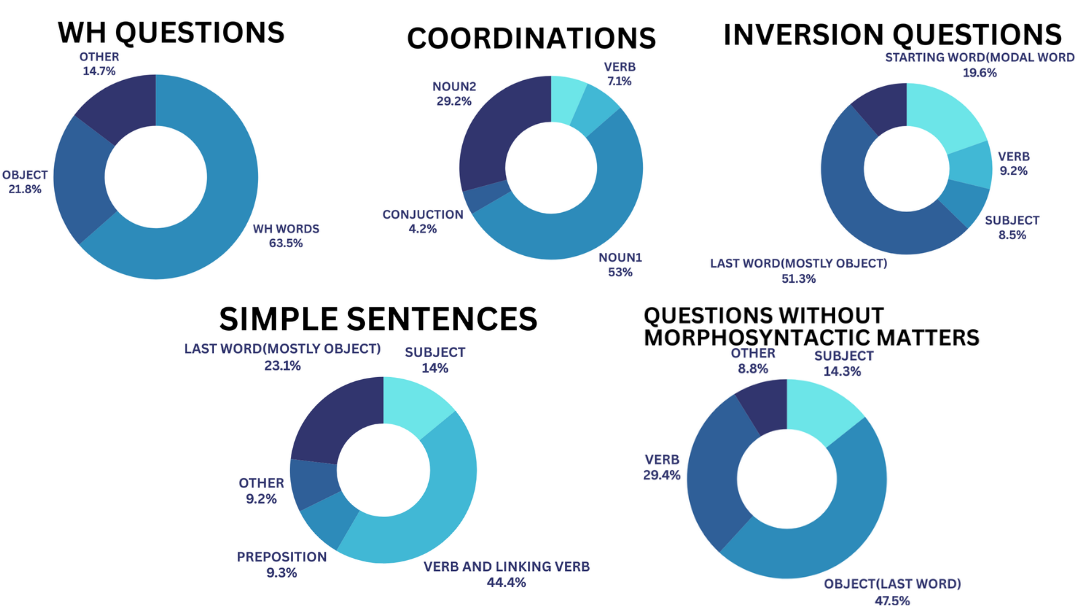}
  \caption{ Analysis of emphasized words - all categories }
  \label{figEW}
\end{figure}

For all utterances of all the five categories, we manually annotated the emphasis words by listening to the entire audio. Based on these annotations, we observed the following: Fig. \ref{figEW}  shows the distribution of emphasis words for the all categories.

The figures reveal similarities in the emphasis on WH-words within the WH-questions category. However, the words that were expected to be emphasized in the other four categories did not appear at the anticipated locations. This led to ambiguous conclusions, as a large number of speech samples in some categories produced different results.

Furthermore, while we attempted to classify emphasized words in Malayalam, we encountered difficulties in accurately identifying and categorizing them. This led to an inability to discern significant differences between the English and Malayalam versions. This limitation highlights the complexity of language perception and processing, especially across different linguistic contexts, and emphasizes the need for further research to develop more refined methods for analyzing language variability and emphasis. As a result, our study did not achieve the expected outcomes in terms of revealing substantial language variability or establishing clear relationships between language features across translations. A deeper analysis is needed in future research to gain better insights into the complexities of language variation and perception across different linguistic contexts.

\section{Conclusion}
In conclusion, our study aimed to analyze prosody and perception in individuals with depression compared to those without, in a language-agnostic manner. To achieve this, we collected data from 132 Malayalam bilingual speakers, annotating their depression states along with emphasized words in their entire speech data. Through careful experimentation and analysis, we demonstrated that our CNN model is a robust and effective tool for automated depression assessment, achieving promising results in detecting depressive symptoms from speech recordings in both English and Malayalam. While our study has made significant advances in machine learning techniques and linguistic research, further investigation is needed to refine our methodologies and deepen our understanding of language variability and mental health assessment. By bridging the gap between machine learning and linguistics, our research contributes to the broader goal of leveraging technology to enhance mental health evaluation and support in multicultural contexts, paving the way for more effective communication and intervention strategies in the future using our bilingual English and Malayalam speech dataset.

\bibliographystyle{unsrt}

\vspace{12pt}
\end{document}